\documentclass[a4paper, conference]{IEEEtran}
\IEEEoverridecommandlockouts

\usepackage{cite}
\usepackage{amsmath,amssymb,amsfonts}
\usepackage{mathabx}
\usepackage{algorithmic}
\usepackage[ruled,vlined]{algorithm2e}
\usepackage{graphicx}
\usepackage{subfig}
\usepackage{array}
\graphicspath{{./}}
\usepackage{hyperref}
\usepackage{textcomp}
\usepackage{xcolor}
\usepackage{comment}
\usepackage{float}
\usepackage{multicol}
\usepackage{multirow}
\usepackage{ragged2e}
\usepackage{enumerate}
\usepackage{fancyhdr}
\newcolumntype{C}{>{\centering\arraybackslash}m{3cm}}
\def\BibTeX{{\rm B\kern-.05em{\sc i\kern-.025em b}\kern-.08em
    T\kern-.1667em\lower.7ex\hbox{E}\kern-.125emX}}
\begin{document}

\title{Self-Supervised Representation Learning for Detection of ACL Tear Injury in Knee MR Videos\\}


\author{\IEEEauthorblockN{ Siladittya Manna}
\IEEEauthorblockA{\textit{CVPRU, Indian Statistical Institute}\\
Kolkata, India \\
siladittya\_r@isical.ac.in}
\and
\IEEEauthorblockN{Saumik Bhattacharya}
\IEEEauthorblockA{\textit{E\&ECE, Indian Institute of Technology}\\
Kharagpur, India \\
saumik@ece.iitkgp.ac.in}
\and
\IEEEauthorblockN{Umapada Pal}
\IEEEauthorblockA{\textit{CVPRU, Indian Statistical Institute}\\
Kolkata, India \\
umapada@isical.ac.in}

}

\maketitle

\begin{abstract}
The success of deep learning based models for computer vision applications requires large scale human annotated data which are often expensive to generate. Self-supervised learning, a subset of unsupervised learning, handles this problem by learning meaningful features from unlabeled image or video data. In this paper, we propose a self-supervised learning approach to learn transferable features from MR video clips by enforcing the model to learn anatomical features. The pretext task models are designed to predict the correct ordering of the jumbled image patches that the MR video frames are divided into. To the best of our knowledge, none of the supervised learning models performing injury classification task from MR video provide any explanation for the decisions made by the models and hence makes our work the first of its kind on MR video data. Experiments on the pretext task show that this proposed approach enables the model to learn spatial context invariant features which help for reliable and explainable performance in downstream tasks like classification of Anterior Cruciate Ligament tear injury from knee MRI. The efficiency of the novel Convolutional Neural Network proposed in this paper is reflected in the experimental results obtained in the downstream task.
\end{abstract}


\section{Introduction}

Deep learning techniques have displayed great success in computer vision tasks like object detection, tracking, segmentation, etc. \cite{AlexNet,VGGNet,DenseCNN,EfficientNet,Inceptionv4}. These deep learning models are trained on datasets containing several gigabytes of human annotated data. Annotating such huge amount of data is time consuming and requires expert domain knowledge. Several attempts have been made to devise techniques to help the machine learning models learn good representation of the underlying data distribution without the availability of large amount of annotated data. Recent advances made in this regard include transfer learning \cite{transferlearning}, semi-supervised learning \cite{semisup,ImprovabilitySemiSuper}, weakly-supervised learning \cite{DeepWeaklySuper}, etc.

In this paper, we have concentrated our efforts on self-supervised representation learning, which is a subclass of unsupervised learning. Self-supervised learning can be used to learn meaningful feature representations from spatial, temporal or spatio-temporal data without the help of human supervision. This objective is generally achieved by solving various pretext tasks, like image inpainting \cite{Pathak2016ContextEFInpainting}, solving jigsaw puzzles \cite{Kim2018LearningIRJigsaw,UnsupervisedLOJIGSAW,Wei2019IterativeRWJigsaw,VideoJigsaw}, temporal order correction \cite{Buckchash2019SustainedSPTOV,ElNouby2019SkipClipSS,Xu2019SelfSupervisedSLVCOP,Siar2020UnsupervisedLOTOP,Misra2016ShuffleALTOP,Fernando2017SelfSupervisedVROOO}, geometric transformation prediction \cite{Jing2018SelfsupervisedSF2,Jing2018SelfSupervisedSF1,Yamaguchi2019MultiplePF}, etc. The pretext tasks and the associated labels are generally defined depending on the nature of the data. The objective of the pretext task is to extract explainable and transferable representations that can be useful in solving a \textit{downstream} tasks, such as, object detection, tracking, semantic segmentation etc. However, in medical image analysis, applications of self-supervised learning methods are limited. Jiao et al. (2018) \cite{Jiao2020SelfSupervisedRLUS} applied a combination of temporal order correction and geometric transformation prediction methods for standard plane detection in fetal ultrasound videos.\\
\indent The objective of this paper is to propose a self-supervised representation learning method to learn features from MR videos of knee without human annotations. These features are used to reliably detect ACL Tear injuries sustained in the knee of a human in the downstream task. The pretext task in our method attempts to solve a jigsaw puzzle and learns meaningful visual representations by solving it. We have shown with rigorous experimental evidences that this method helps the pretext models to learn spatial context-invariant features in MR video clips, unlike previous works where the features learnt by the pretext models are covariant to the transformations applied \cite{Kim2018LearningIRJigsaw, Pathak2016ContextEFInpainting, UnsupervisedLOJIGSAW, Gidaris2018UnsupervisedRLImageRot}.

The contributions of this work are as follow:
\begin{itemize}
    \item We propose a novel Convolutional Neural Network architecture for efficiently solving jigsaw puzzle as pretext task. This model can be trained from scratch to learn explainable visual representational features.
    \item We also propose an unique \textit{Divide-and-Teach} strategy to train the model for the downstream task in case of GPU memory constraint. This strategy also enables the model to learn temporally independent features.
    \item Our work is demonstrated to be effective in extracting explainable and transferable context invariant features as evident from results obtained in the downstream task.
\end{itemize}

\section{Methodology}
\label{sec: Methodology}

\noindent
In this paper, the goal is to learn feature representation of the spatio-temporal information available from the MR videos. We achieve this goal by devising a novel CNN architecture, which predicts the order in which the patches. The arrangements are chosen using Algorithm \ref{alg:alg1} and the patches are arranged according to the chosen arrangement using Algorithm \ref{alg:alg2}. In the following subsections, we focus on designing the pretext algorithm and subsequently the downstream algorithm for detecting ACL tear injury from knee magnetic resonance videos.

\subsection{Pretext Task Algorithm}
\label{subsec: pretextalgo}
\begin{figure}[ht]
\label{fig:eximgs}
\centering
\begin{tabular}{c c}
\subfloat[]{\includegraphics[scale=0.3]{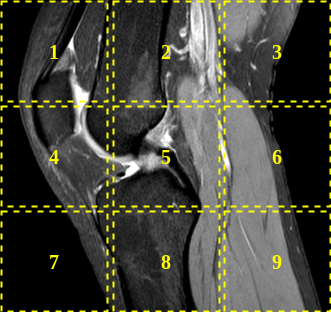} \label{fig:eximga}} & 
\subfloat[]{\includegraphics[scale = 0.265]{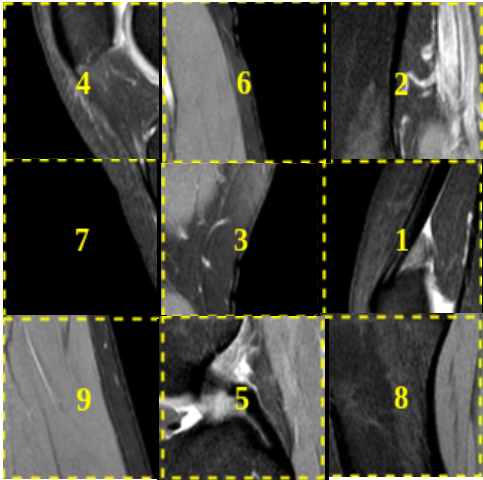} \label{fig:eximgb}}\\
\end{tabular}
\caption{(a) Image showing the numbering of the patches in an ordered frame. (b) Image showing the patches after being arranged using Algorithm \ref{alg:alg2}}
\end{figure}

The pretext task in our method is similar to jigsaw puzzle solving strategy. In this learning strategy, we divide a randomly chosen frame of a MR video clip into $N$ square patches of dimension $\lfloor \frac{L}{\sqrt{N}} \rfloor \times \lfloor \frac{L}{\sqrt{N}} \rfloor$, where $L$ is the dimension of the square frame, $N$ is the number of patches we want to divide the frame into, and $\lfloor x \rfloor$ equals the nearest integer less than or equal to $x$. Dividing the frame into $N$ patches gives $N!$ ways to jumble the patches. For $N = 9$, we have $9! = 3,62,880$ rearrangements. Let us denote the set of all arrangements as $\mathcal{J}$. Also, let the rearrangement, applying which the frame remains ordered, as in Fig. \ref{fig:eximga}, be denoted by $\tau_{0}$. For our work, $L = 256$ and $N = 9$, thus $\lfloor \frac{L}{\sqrt{N}} \rfloor = 86$.

Since solving a classification task with such a large number of classes would require a huge amount of data and computational time, we choose a subset $\mathcal{A} \subset \mathcal{J}$ by following the Algorithm \ref{alg:alg1}.
\begin{algorithm}
\SetAlgoLined
\KwResult{$\mathcal{A}$ : Set of Arrangements}
 \textbf{Given }
 
 $\mathcal{J}$ : Set of all possible arrangements\\
 $\mathcal{U}_{A}$ : A is a sample drawn from uniform distribution $\mathcal{U}$
 $\mathcal{C}$ : Number of classes in the pretext task
 
 \textbf{Initialize }
 $\mathcal{A'} = \mathcal{A} = \{\,\tau_0\,\}$;\\
 
 \For{i = 1 : 9! - 1}{
  \If{$hammingDist(a, \mathcal{J}[i]) > 4 \forall a \in \mathcal{A'}$}{
   $\mathcal{A'} = \mathcal{A'} \,\bigcup \,\{\mathcal{J}[i]\}$;
   }
 }
 \For{i = 1 : $\mathcal{C}$ - 1}{
 $\mathcal{A} = \mathcal{A} \,\bigcup \, \mathcal{U}_{A}[\mathcal{A'}]$
 }
 \caption{SETARR : How to choose the set of arrangements}
 \label{alg:alg1}
\end{algorithm}
which describes the steps we use to choose the permuted rearrangement  orders to be included in $\mathcal{A}$. We initialize the set of arrangements with the ordered arrangement $[1,2,3,4,5,6,7,8,9] (\tau_0)$ and choose the threshold of hamming distance as $\lfloor \frac{N}{2} \rfloor$. The hamming distance between two permutations is defined as the number of positions in which they differ. In our experiments, we consider $N = 9$, thus the threshold of hamming distance equates to $4$. We progressively keep on adding elements from the set $\mathcal{J}$ if the hamming distance from the all elements in the set $\mathcal{A'}$ is more than 4. This algorithm ensures that the elements in the chosen set are neither too close nor too far from other elements in the permutation space. This maintains a balance in the difficulty of the pretext classification task.\\
\indent
Running Algorithm \ref{alg:alg1} on $\mathcal{J}$ resulted in a subset of 1887 permutations. We adopted a uniform random sampling without replacement strategy according to an uniform distrbution $\mathcal{U}$, to get the reduced set $\mathcal{A}$ arrangements from the chosen 1887 arrangements. It should be noted that the number of arrangements, $\mathcal{A}$ is equal to the number of classes $\mathcal{C}$ in the pretext classification task.

We chose the augmentation from a finite set $\mathcal{G}$, which can be expressed as a Cartesian product of four finite sets $\mathcal{R}$, $\mathcal{T}_{x}$, $\mathcal{T}_y$ and $\mathcal{S}$, i.e., 
$\mathcal{G} = \mathcal{R} \bigtimes \mathcal{T}_{x} \bigtimes \mathcal{T}_{y} \bigtimes \mathcal{S}$, where $\mathcal{R}=\{-15^{\circ},0^{\circ},15^{\circ}\}$, $\mathcal{T}_{x}=\mathcal{T}_{y}=\{- \lfloor 0.1L_{p} \rfloor,0,\lfloor 0.1L_{p} \rfloor\}$ and $\mathcal{S}= \{1,1.2\}$. Here $\mathcal{R}, \mathcal{T}_{x}, \mathcal{T}_{y}, \mathcal{S}$ denote the finite sets of angles of rotation in degrees, magnitude of translation along $x$-axis and $y$-axis in pixels and scale factors, respectively. $L_{p}$ denotes the dimension of each side of a square patch obtained after applying Algorithm \ref{alg:alg2}.

To obtain the jumbled patches (Fig. \ref{fig:eximgb}) and the pretext labels, we apply Algorithm \ref{alg:alg2} to the frames randomly sampled from each MR video. Firstly, each frame $\mathcal{F}$ is partitioned into $N$ parts, each of dimension $\lfloor \frac{L}{\sqrt{N}} \rfloor$. Augmentation $g$ is obtained by uniformly sampling an element from the finite set $\mathcal{G}$ and applied to each partition denoted by $\textbf{map}_{\mathcal{I}}(g,\mathcal{P}')$. Then, the reference point ($ref_{x}$, $ref_{y}$) for each patch is obtained by uniformly sampling values from the range [0, $\lfloor \frac{L}{\sqrt{N}} \rfloor - 64$]. A patch of dimension $64 \times 64$ is cropped from the larger partition $\mathcal{P}'$ of dimensions $\lfloor \frac{L}{\sqrt{N}} \rfloor \times \lfloor \frac{L}{\sqrt{N}} \rfloor$ with the reference point ($ref_{x}, ref_{y}$) as its origin. Finally, the patches are arranged according to an arrangement $\mathcal{T}$ drawn from the set $\mathcal{A}$ according to an uniform distribution over the set, to get the jumbled patches $\mathcal{P}_A$ (using $\textbf{map}_{\mathcal{I}}(\mathcal{T},\mathcal{P}_A)$). $\mathcal{P}_A$ is the input to the pretext model and the arrangement $\mathcal{T}$ is the corresponding pretext ground truth label. It should be mentioned here that $L_{p} \neq \lfloor \frac{L}{\sqrt{N}} \rfloor$. In our experiments, we set $\lfloor \frac{L}{\sqrt{N}} \rfloor = 85$. 

\begin{algorithm}[h]
\SetAlgoLined
\KwResult{$\mathcal{P}_A$ : Jumbled Patches from a frame $\mathcal{F}$}
 \textbf{Given}
 
 $A$ : Set of rearrangements\\
 $\mathcal{G}$ : Set of geometric transformations\\
 $\mathcal{U}_z[\cdot]$ : z is a sample drawn from uniform distribution $\mathcal{U}$ over any set\\
 $\mathcal{T}$ : an arrangement to be applied on the patches and sampled uniformly from the set $\mathcal{A}$\\
 $\textbf{map}_{\mathcal{I}}(\cdot)$ : a function which denotes a given arrangement or augmentation being applied to an image (or patch)\\
 
 \textbf{Initialize} 
 
 $\mathcal{F}$ : a random frame from a MR video sample\\
 $\mathcal{P}_A = \{\,\}$\\
 $\mathcal{L'} = \lfloor \frac{L}{\sqrt{N}} \rfloor$\\
 $row = col = ref_{x} = ref_{y} = 0$
 
 \For {i=1:9}{
 $row = \lfloor \frac{i}{\sqrt{N}} \rfloor$ \\
 $col = i \mod \sqrt{N}$\\
 $\mathcal{P'} =  \mathcal{F}[row. \mathcal{L'}  : (row+1) \mathcal{L'} , col. \mathcal{L'} : (col+1) \mathcal{L'}]$\\
 $g = \mathcal{U}_{g}[\mathcal{G}]$\\
 $\mathcal{P'} = \textbf{map}_{\mathcal{I}}(g,P')$\\
 $ref_{x} = \mathcal{U}_{x}[0,\mathcal{L'}-64]$\\
 $ref_{y} = \mathcal{U}_{y}[0,\mathcal{L'}-64]$\\
 $\mathcal{P'} = \mathcal{P'}[ref_{x} : ref_{x}+64, ref_{y} : ref_{y}+64]$\\
 $\mathcal{P}_A = \mathcal{P}_A \bigcup \mathcal{P'}$\\
 }
 
 $\mathcal{T} : \mathcal{U}_{\tau}[\mathcal{A}]$\\
 $\mathcal{P}_A = \textbf{map}_{\mathcal{I}}(\mathcal{T},\mathcal{P}_A)$
 \caption{JUMPAT : How to obtain the jumbled patches}
 \label{alg:alg2}
\end{algorithm}

\subsection{Motivation behind proposed architecture}
\label{subsec:issues}
\noindent
Pretext task models are very prone to learning low level signals like void regions, boundary edges and corners, etc. When using the jigsaw puzzle solving strategy without the augmentations, the model tends to learn low level signals similar to the clues that humans often use when solving jigsaw puzzles. The approach we follow in this paper also compels us to take a subset of the large pool of possible rearrangements.

\begin{figure}[!ht]
    \centering
    \includegraphics[width=\linewidth]{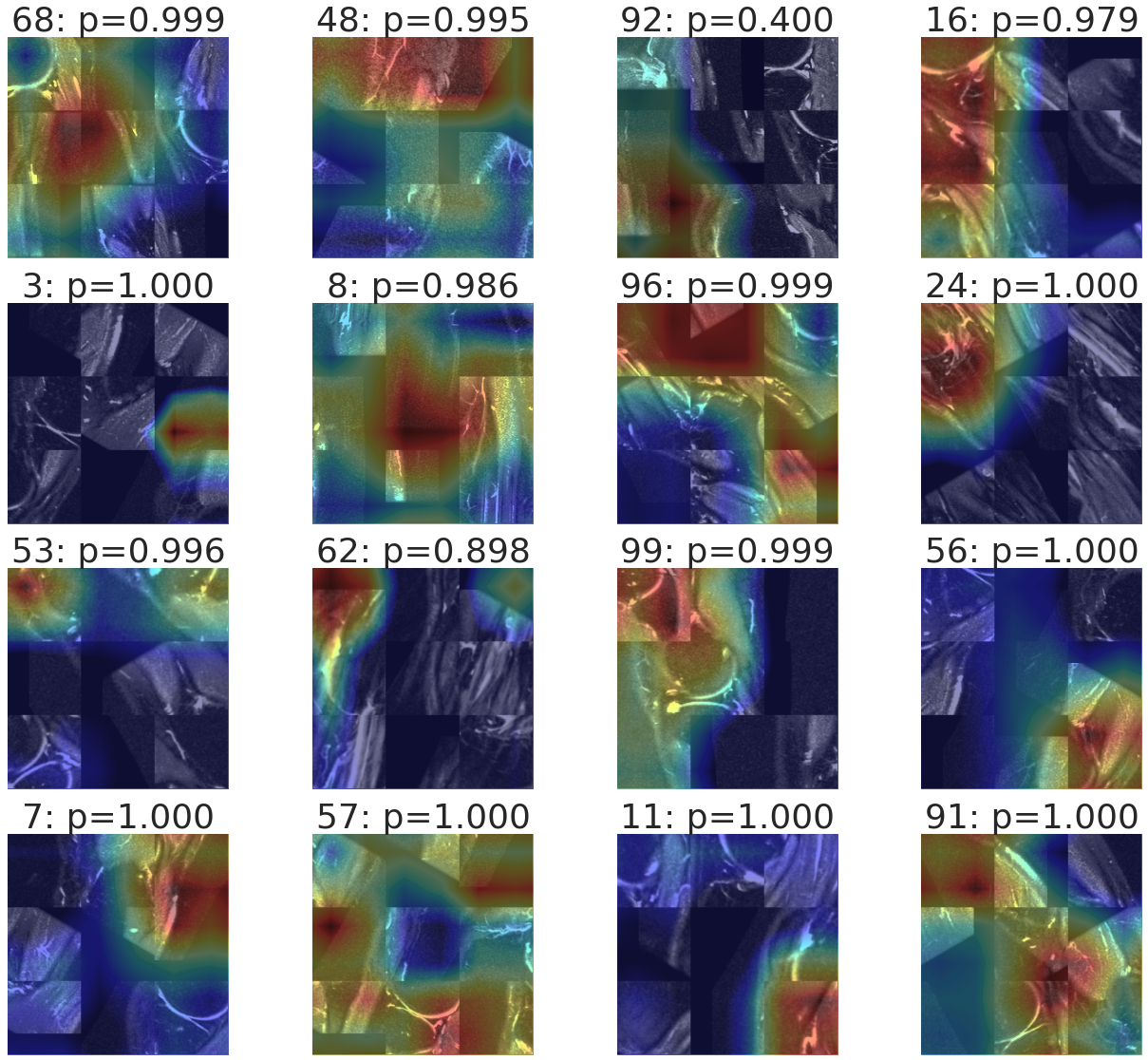}
    \caption{Gradcam output shows the regions (indicated by red) where the model built using pre-trained Inception-ResNet-v2 gains maximum information. It is clearly visible that the maximum attention is on the low level signals as mentioned in Section \ref{subsec:issues}}
    \label{fig:gradcam1}
\end{figure}

In one of our initial models, we used a single Inception-ResNet-v2 network pre-trained on ImageNet, to detect the arrangements. The input was all the 9 patches put together like in Fig. \ref{fig:eximgb}. After analysing the learned feature maps, we observed that the model used the low level signals like boundary corners and edges and discontinuities between patches to learn discriminative features. This tendency of the model to learn features without proper generalization of the loss surface prevents it from learning meaningful context invariant visual representational features. Fig. \ref{fig:gradcam1} shows the gradient class activation mapping \cite{GradCAM} outputs of the aforementioned model along with the ground truth label and the probability of prediction.

\subsection{Pretext Task Model Architecture}
\label{subsec:pretextarch}
\begin{figure*}
    \centering
    \includegraphics[scale = 0.08]{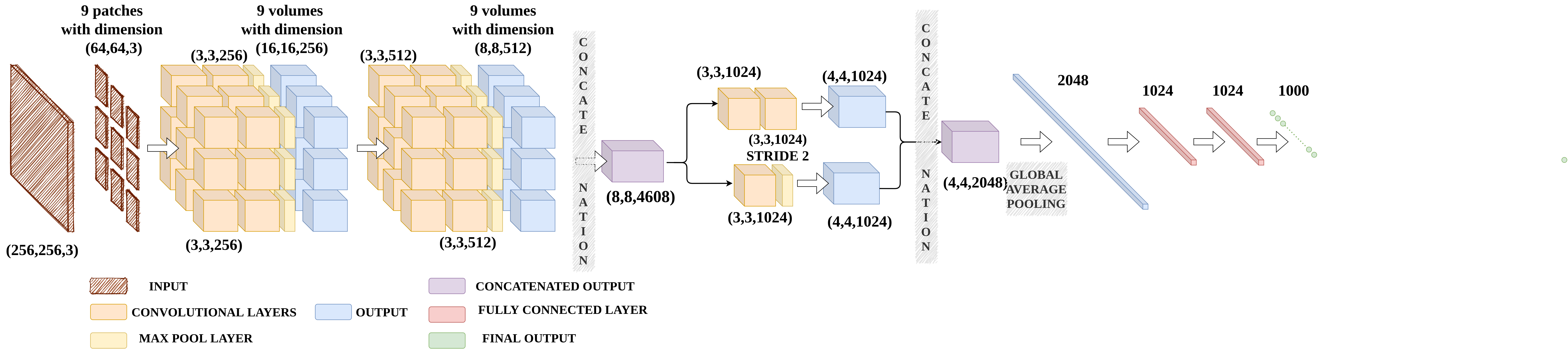}
    \caption{Proposed network model for Pretext Task.}
    \label{fig:pretextmodelarch}
\end{figure*}

In this paper, we have used a semi-parallel architecture for our pretext tasks, where we predict the order in which the patches are arranged. We call this architecture JPOPNet (JPOP stands for Jumbled Patch Order Prediction) and is shown in Fig. \ref{fig:pretextmodelarch}. The results presented in the previous section show the reason behind the adoption of a semi-parallel architecture in this paper. We feed each of the 9 patches in the input into one of the 9 parallel convolutional channels. Each convolutional channel is made up of 2 Convolutional blocks. Each convolutional block consists of two convolutional layers followed by a maxpooling layer. The number of filters of both the convolutional layers, in the two convolutional blocks are 256 and 512, respectively. The maxpooling layer has a pool window of dimensions $2 \times 2$ and a stride of $2$. 

The output from all the 9 channels are then concatenated to get a output volume of dimension $16 \times 16 \times 4608$. This output volume is then convoluted with a convolutional layer with filters $3 \times 3 \times 2048$ to reduce the dimensionality and gives an output of dimensions $16 \times 16 \times 2048$, which is then fed into two separate branches. The first branch is a convolutional block, which consists of two convolutional layers with 1024 filters, with only the second layer having stride 2, thereby causing the spatial dimension of the output to be reduced to half of its input. The second branch contains a convolution layer with 1024 filters with kernel size $3 \times 3$ and followed by a maxpooling layer which reduces the dimensions to half. The outputs from the two channels are again concatenated to form an output volume of dimension $8 \times 8 \times 2048$. Global average pooling is applied to this output of dimension $8 \times 8 \times 2048$ to obtain an output of dimension $2048$, which is then fed into a network of two fully connected layers of dimension 1024. The second fully connected layer is connected to the output consisting of $\mathcal{C}$ nodes, where $\mathcal{C}$ is the number of classes.

\begin{figure*}
\begin{minipage}{\textwidth}
    \centering
    \includegraphics[scale = 0.066]{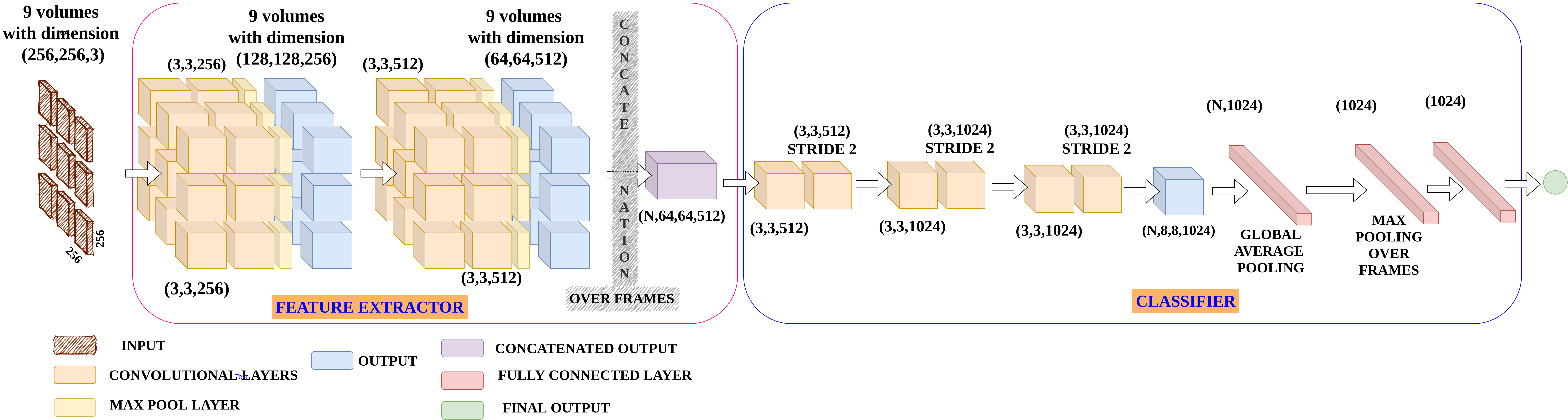}
    \caption{Network model used for Downstream Task}
    \label{fig:dsmodelarch}
\end{minipage}
\end{figure*}

\subsection{Downstream Task Algorithm}
\label{subsec:dsalgo}
In this paper, the objective of the downstream task is to predict whether the knee has sustained injury to the Anterior Cruciate Ligament or not. 

\begin{algorithm}[ht]
\SetAlgoLined
\KwResult{$\textbf{I}$ : Input}
 \textbf{Initialize}\\ $\textbf{I} = [\,]$;\\$\textbf{start\_index} = 0$;\\
 $\textbf{end\_index} = 0$;\\
 \textbf{Given}\\ $\textbf{F}$ : All frames in the MR videosclip\\\textbf{N} : Number of frames\\
 \For{i = 1 : 9}{
  $n = \lceil \frac{N}{9-i+1} \rceil$\\
  end\_index = end\_index + n\\
  Append \textbf{F}\,[ start\_index \,:\, end\_index ] to \textbf{I}\\
  
  start\_index = end\_index\\
 }
 \caption{DIVFRAM : How to divide the frames temporally}
 \label{alg:alg3}
\end{algorithm}

For performing the downstream task, we construct a model (Fig. \ref{fig:dsmodelarch}) consisting of two parts, \textit{feature extractor} and \textit{discriminator}. From the pretext model, the 9 branches with 4 convolutional layers each acts as the feature extractors. We also devise an unique \textit{Divide-and-Teach} training methodology. Since the frames were uniformly sampled from each MR video, each convolutional layer is capable of extracting useful features from the frames, irrespective of the temporal position of the frame in the MR video. We divide $|F|$ frames into 9 parts before feeding to the 9 channels of the CNN, $|F|$ being the total number of frames in the MR video. After the respective outputs are obtained from each channel, we concatenate the outputs over the frames to obtain an output of dimension $|F| \times 64 \times 64 \times 512$, which is then fed into the classifier to obtain the predictions.\\

\subsection{Downstream Model Architecture}
\label{subsec:dsarch}
The downstream model consists of two parts: feature extractor and the discriminator, as shown in Fig. \ref{fig:dsmodelarch}. The feature extractor is made up of the 9 parallel branches of the pretext model. The output from the 9 branches are concatenated to form an output of dimensions $(64 \times 64 \times 512)$. The output obtained from the feature extractor is fed to the discriminator.\\
\indent The classifier consists of three convolutional blocks, each containing two convolutional layers. Both the layers in each convolutional block has filter size $3 \times 3$ but only the second convolutional layer has stride 2. This reduces the dimensions to half without the use of maxpooling layers. The three convolutional layers result in an output of shape $|F| \times 8 \times 8 \times 1024$. We then apply Global Average Pooling to the output, followed by maxpooling over frames. This gives an output of dimension $1024$, which is then fed into a network of two fully connected layers, each containing $1024$ nodes. The output from this layer is finally fed into the output node. The downstream task is a binary classification task, hence sigmoid activation is applied on the output node to obtain predictions in the 0 to 1 range. 

\section{Experiment and Results}
\subsection{Dataset}
In our experiments, we use the MRNet \cite{Bien2018DeeplearningassistedDF} dataset as our reference dataset. The MRNet dataset contains 1370 knee MR video clips in total. Out of 1370 clips, 1130 MR video clips are included in the training set and 120 MR video clips are considered as the tuning or validation set. The rest 120 are used for external validation. Out of the 1,130 training examples, only 208 videos contain ACL tear. It is evident that the dataset we are using for this work is highly imabalanced. This gives us an opportunity to explore the effects of self-supervised learning techniques on imbalanced datasets.

\subsection{Pretext Task Experimental Details}
\label{subsec:pretextexp}
The model was trained with data produced following the algorithm discussed in Section \ref{subsec: pretextalgo}. We optimized the categorical cross-entropy loss of the model using RMSprop optimizer with an initial learning rate of $10^{-4}$ decayed at the rate of 0.95 per epoch. We used a batch size of 32 during both training and validation stages. Since, our ultimate goal is to extract features from frames which are not jumbled, it seemed logical to tune the network only on the ordered frames. The pretext model was trained entirely from scratch on a NVIDIA RTX 2080Ti 11GB GPU. The training was stopped when the validation accuracy flattened.

\subsection{Pretext Task Results}
\label{subsec:pretextres}
\begin{figure}[ht]
\centering
\begin{tabular}{cc}
    \subfloat[]{\includegraphics[scale = 0.45]{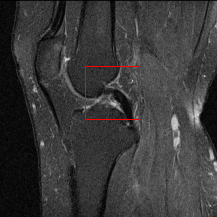} \label{fig:5a}} & 
    \subfloat[]{\includegraphics[scale = 0.45]{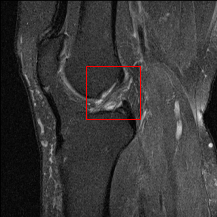} \label{fig:5b}}\\
    
    \subfloat[]{\includegraphics[scale = 1.35]{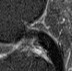} \label{fig:5c}}&
    \subfloat[]{\includegraphics[scale = 1.35]{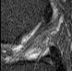} \label{fig:5d}}\\
\end{tabular}
\caption{Region of Interest for ACL tear detection. Image (c) and (d) showing the enlarged view of the ROIs marked in red in the image (a) and (b), respectively. The image in Fig. \ref{fig:5a} (also \ref{fig:5c}) and \ref{fig:5b} (also \ref{fig:5d}) are examples of a torn ACL and a uninjured ACL, respectively.}
\label{fig:acleg}
\end{figure}

In this subsection, we have presented the results of ACL tear injury detection from Knee MR videos. In Fig. \ref{fig:acleg}, we can see the region which needs to be focused on.. To analyze the generalization and feature learning capacity of the model, we train with 500 and 1000 random permutations chosen according to as Algorithm \ref{alg:alg1}. As shown in Table \ref{tab:accuracies}, even after increasing the number of permutations, the proposed model performs well on the pretext tasks, thereby justifying the capability of the model in learning meaningful features to efficiently distinguish between such large number of equi-spaced permutations of the image patches.

\begin{table}[ht]
\centering
\bgroup
\def\arraystretch{1.5}
\caption{Pretext Task Experimental Results}
\label{tab:accuracies}
\begin{tabular}{|c|c|c|}
\hline
 \multirow{2}{2cm}{\centering No. of permutations} & \multirow{2}{2cm}{\centering No. of parameters} & \multirow{2}{1.5cm}{\centering Validation Accuracy}  \\
 {} & {} & {}\\\hline\hline
 500 & 173 Million  & 96.4\% \\\hline
 1000& 173.5 Million & 93.5\%  \\\hline
\end{tabular}
\egroup
\end{table}

\subsection{Downstream Task Experimental Details}
In the downstream task, due to memory constraint on the NVIDIA RTX 2080Ti, we limited the number of frames to $min(|F|,36)$, where $|F|$ is the number of frames in the MR video clips. If the number of frames in any MR video clip is more than 36, we use uniform random sampling to select 36 frames from $|F|$ number of frames. This strategy helps the model deal with missing frames and also temporally sparse data. Also, we kept the batch size limited to 1. The downstream model was trained by optimizing the binary cross-entropy loss of the model using Adam optimizer with an initial learning rate of $10^{-5}$. Since the dataset is highly imbalanced, we used oversampling to balance the dataset before training our model. The number of positive ACL tear injury samples in the MRNet dataset is 208 and the number of negative samples is 922. We oversampled the minority class to 922. This oversampled dataset was then used to train the downstream model. During the validation stage also, we chose $min(|F|,36)$ number of frames and then partitioned the frames into 9 parts using Algorithm \ref{alg:alg3}. Apart from the \textit{Divide-and-Teach} training strategy mentioned in Sec. \ref{subsec:dsalgo}, data augmentations like random rotation, translation and scaling were also applied on each frame during training.

\subsection{Downstream Task Results}

\begin{figure}[ht]
\centering
\begin{tabular}{cc}
    \subfloat[]{\includegraphics[scale = 0.45]{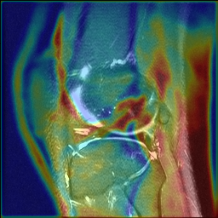}} & 
    \subfloat[]{\includegraphics[scale = 0.45]{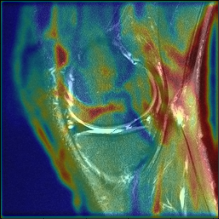}}\\

    \subfloat[]{\includegraphics[scale = 1.38]{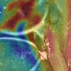}} &
    \subfloat[]{\includegraphics[scale = 1.29]{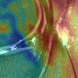}}\\
\end{tabular}
\caption{Gradient Activation Class Mappings on the $256 \times 256$ frames show that the  downstream task model focuses on the ACL. image (c) and (d) are the zoomed in versions of the images in (a) and (b),respectively like in Fig. \ref{fig:acleg}}
\label{fig:dsgradcam}
\end{figure}

In the downstream task, we gradually increased the number of parameters by adding different layers. As the number of parameters increases, the models' capability of approximating the function mapping from the input space to the output space also increases. It can be observed from the results presented in Table \ref{tab:dsres} that increasing the number of parameters boosts the performance, even when the positive samples are under-represented in the pretext task. The detailed ablation studies are described in Sec. \ref{subsec:ablstu}. For the ACL tear detection task, the best results were obtained using our final model with 77 million parameters. It achieved an accuracy of 76.62\% (95\% CI 74.50, 78.83) on the validation set and an AUC score of 0.8481 (95\% CI 0.8284, 0.8651).\\
\indent From the gradient class activation mappings in Fig. \ref{fig:dsgradcam}, it can be seen that the downstream model focuses on the desired region. Though, there are some models that perform the classification task of injury from MR video frames, to the best of our knowledge, none of them provides any insight or explanation for the decision made by their models.

\subsection{Ablation Studies}
\label{subsec:ablstu}

To optimize our model architecture, we built multiple models by changing the different hyperparameters associated with the model. Among all the variants, the model shown in Fig. \ref{fig:dsmodelarch} corresponds to the final model which performed the best in the downstream task. In a variant (Model-1), a maxpooling layer was introduced instead of the first convolutional block in the Discriminator and the two convolutional layers in the second convolutional block contained 512 filters each. Also, only one fully connected layer was used in Model-1. In the second variation (Model-2), we increased the capacity by adding another fully connected layer with 1024 nodes and increasing the number of filters of the convolutional layers in the second convolutional block to 1024. The maxpooling layer in the classifier of Model-1 remains unchanged in Model-2. In the best performing model (Proposed, shown in Fig. \ref{fig:dsmodelarch}), we replaced the maxpooling layer in the discriminator by a convolutional block containing two convolutional layers with 512 filters each and only the second layer have a stride of $2$. The performance results of all the three models have been shown in Table \ref{tab:dsres}.

\begin{table}[!h]
    \centering
    \bgroup
    \def\arraystretch{1.5}
    \caption{Ablation study on downstream task for detection of ACL injury}
    \begin{tabular}{|c|c|c|c|}
    \hline
        \multirow{2}{1.3cm}{\centering Model} & 
        \multirow{2}{1.5cm}{\centering Number of parameters} & 
        \multirow{2}{2cm}{\centering Accuracy (5\%-95\% CI)} & 
        \multirow{2}{2cm}{\centering AUC (5\%-95\% CI)}\\
        {} & {} & {} & {}\\
        \hline \hline
        \multirow{2}{1.3cm}{\centering Proposed} & 
        \multirow{2}{1.5cm}{\centering 77 Million} & 
        \multirow{2}{2cm}{\centering 76.62 (74.5-78.83)} & 
        \multirow{2}{2cm}{\centering 0.848 (0.828-0.865)}\\
        {} & {} & {} & {}\\
         \hline
         \multirow{2}{1.3cm}{\centering Model-2} & 
        \multirow{2}{1.5cm}{\centering 75 Million} & 
        \multirow{2}{2cm}{\centering 73.4 (71.0-75.6)} & 
        \multirow{2}{2cm}{\centering 0.834 (0.812-0.850)}\\
        {} & {} & {} & {}\\
         \hline
         \multirow{2}{1.3cm}{\centering Model-1} & 
        \multirow{2}{1.5cm}{\centering 72 Million} & 
        \multirow{2}{2cm}{\centering 71.7 (70.2-72.9)} & 
        \multirow{2}{2cm}{\centering 0.813 (0.797-0.829)}\\
        {} & {} & {} & {}\\
         \hline
    \end{tabular}
    \label{tab:dsres}
    \egroup
\end{table}

\subsection{Effects of Class Imbalance}

The pretext and the downstream task, both contribute to the ultimate objective of detection of ACL injury from knee MR videos. The motivation of our work is to build a pretext model, capable of learning spatial context invariant visual representational features. The results presented in Table \ref{tab:ablation} show that in case of an imbalanced dataset, the features of the majority class receive more weightage than the minority class in the pretext task. Every sample in the training set is chosen exactly once when preparing the pretext training samples. Thus, for each pretext label, there are more samples from the majority class than from the minority class.

When the oversampled dataset is used to train the model in the pretext task, equal number of samples from both classes are selected for preparing the training samples. Thus, the features from both the original classes are learnt with equal weightage. The downstream model showed an increase in the \textit{True Positive Rate} and a reduction in \textit{Type 2 error}. However, \textit{Type 1 error} increased slightly, subsequently lowering \textit{True Negative Rate}.

\begin{table}[!h]
    \centering
    \bgroup
    \def\arraystretch{1.5}
    \caption{Ablation Study on the effects of class imbalance on the task of detection of ACL tear injury}
    \label{tab:ablation}
    \begin{tabular}{|c|c|c|}
    \hline
        \multirow{2}{2cm}{\centering Model} & 
        \multirow{2}{2cm}{\centering Accuracy (5\%-95\% CI)} & 
        \multirow{2}{2cm}{\centering AUC (5\%-95\% CI)}\\
        {} & {} & {}\\
        \hline \hline
         \multirow{2}{2cm}{\centering without oversampling}&
         \multirow{2}{2cm}{\centering 76.62 (74.5-78.63)} & 
        \multirow{2}{2cm}{\centering 0.848 (0.828-0.865)}\\
        {} & {} & {}\\
         \hline
         \multirow{2}{2cm}{\centering with oversampling}&
         \multirow{2}{2cm}{\centering 76.72 (74.9-78.70)} & 
        \multirow{2}{2cm}{\centering 0.848 (0.826-0.87)}\\
        {} & {} & {}\\
         \hline
    \end{tabular}
    \egroup
\end{table}

\subsection{Comparison with Supervised Methods}
To compare our method with supervised learning techniques we present the results of the MRNet \cite{Bien2018DeeplearningassistedDF} model on the same dataset. Apart from limiting the number o frames to a maximum of 36, the MRNet \cite{Bien2018DeeplearningassistedDF} was trained using the original conditions. 

\begin{table}[h]
    \centering
    \caption{Comparison with supervised learning method}
    \begin{tabular}{|c|c|c|}
        \hline
        \multirow{2}{1.5cm}{Method} & \multicolumn{2}{c|}{ACL Tear} \\ \cline{2-3}
        {} & Accuracy (\%) & AUC \\ \hline \hline
        MRNet \cite{Bien2018DeeplearningassistedDF} & 86.63 & 0.963 \\ \hline 
        Ours & 76.62 & 0.848 \\ \hline
    \end{tabular}
    \label{tab:supercomp}
\end{table}

\section{Conclusion}

The objective of this work is to explore the capabilities of self-supervised learning algorithms in medical image analysis. It has been shown that our proposed pretext model extracts structural features particularly from the region of interest which can support a downstream task of classification further. The challenges associated with this pretext task are discussed and analyzed thoroughly. However, approaches involving self-supervision depend largely on the quality of the features that the pretext models learn and this shapes the performance of the downstream task. This is the first work of this kind and we look to further explore other techniques which can accommodate different kinds of injuries in a single downstream task by learning more robust and meaningful visual representational features in the pretext tasks. 

{\small
\bibliographystyle{ieeetr} 
\bibliography{root}
}

\end{document}